\documentclass[runningheads]{llncs}
\usepackage[T1]{fontenc}
\usepackage{graphicx}
\usepackage{amsmath,amsfonts}
\usepackage{array}
\usepackage{textcomp}
\usepackage{stfloats}
\usepackage{url}
\usepackage{verbatim}
\usepackage{graphicx}
\usepackage{color}
\usepackage{svg}
\usepackage{pdfpages}
\usepackage{pifont}
\usepackage{threeparttable}
\usepackage[linesnumbered,ruled,vlined]{algorithm2e}
\RequirePackage{letltxmacro}
\usepackage{comment}
\usepackage{paralist}
\LetLtxMacro{\LaTeXtextbf}{\textbf}
\LetLtxMacro{\textbf}{\LaTeXtextbf}
\usepackage[bookmarks=false]{hyperref} 
\usepackage{cite}
\hypersetup{
    colorlinks=true,
    citecolor=blue
}
\usepackage{amsmath,amssymb,amsfonts}
\usepackage{graphicx}
\usepackage{textcomp}
\usepackage{caption}
\usepackage{subcaption}
\usepackage{algpseudocode}
\usepackage{array}
\usepackage{enumitem,lipsum}
\setlist[itemize,enumerate]{leftmargin=*}
\usepackage{etoolbox}
\usepackage{rotating}
\makeatletter
\usepackage{wasysym}
\makeatother
\usepackage{tcolorbox}
\usepackage[table,xcdraw]{xcolor} 
\tcbuselibrary{breakable}
\usepackage{svg}
\usepackage{multirow}
\usepackage{listings}             
\usetikzlibrary{shapes, arrows, positioning}

\newcommand{\ie}{\textit{i.e.}}
\newcommand{\eg}{\textit{e.g.}}

\setlist[enumerate,1]{label=\textbf{\scriptsize\arabic*.}, leftmargin=*}

\newlist{circlelist}{enumerate}{1}
\setlist[circlelist,1]{label=\textbf{\textcircled{\small\arabic*}}, leftmargin=*}

\usepackage{enumitem}
\usepackage{orcidlink}
\usepackage{booktabs}

\usepackage{paralist}

\definecolor{dred}{RGB}{178,34,34}
\definecolor{dgreen}{RGB}{34,110,34}

\usepackage{tikz}
\usetikzlibrary{
arrows.meta,
positioning,
shapes,
fit,
calc
}

\definecolor{brown}{rgb}{0.4,0.2,0.1}

\begin{document}
%

%
%
\title{Robust Reputation-Driven Crowdsourced Federated Learning}

\author{Mouhamed Amine Bouchiha\inst{1}\orcidID{0000-0001-6142-6855} \and Gregory Blanc\inst{1}\orcidID{0000-0001-8150-6617} 
}
\authorrunning{M. Bouchiha et al.}
%
\institute{SAMOVAR, Télécom SudParis, Institut Polytechnique de Paris 
\email{mbouchiha@telecom-sudparis.eu,}
\email{gregory.blanc@telecom-sudparis.eu}
}

\maketitle              
\begin{abstract}
Crowdsourced Federated Learning (CrowdFL) extends traditional federated learning by enabling open and heterogeneous participation through a crowdsourcing paradigm. In this setting, reputation-driven incentive mechanisms are commonly employed to guide worker selection and enhance trustworthiness. While such approaches improve participant reliability, existing frameworks largely overlook the quantification of their robustness against stealthy adversaries, particularly those capable of evading standard detection mechanisms. To fill this gap, this paper proposes R2CFL, a robust reputation-driven CrowdFL framework. R2CFL introduces a robust reputation model coupled with a nearest neighbor mixing (R2-NNM) defense mechanism that links reputation evolution with the filtering of updates during aggregation. The proposed mechanism prevents stealthy attackers from gradually accumulating trust and influencing future tasks. Experimental results demonstrate that R2-NNM matches or surpasses state-of-the-art Byzantine-robust and backdoor defense mechanisms against adaptive attackers. Furthermore, when integrated with existing detect-and-filter defenses, the proposed reputation model faithfully captures the statistical robustness of the underlying defense by producing reputation scores that closely reflect its true positive and false positive characteristics.

\end{abstract}

\section{Introduction}
\label{sec:intro}

Federated learning (FL)~\cite{mcmahan2017communication} enables distributed devices to collaboratively train machine learning models while keeping training data local to improve privacy. Its applicability has expanded to large-scale and dynamic environments where centralized data collection is impractical~\cite{flsrv1}. Crowdsourced Federated Learning (CrowdFL) extends FL by allowing open and heterogeneous participants to contribute to the training process in a flexible and scalable manner~\cite{feng2022crowdfl}. In such settings, participants (workers) may join or leave dynamically, and their reliability and data quality can vary significantly. To address this, reputation-driven incentive mechanisms are commonly employed to guide participant selection, encourage honest behavior, and improve overall system performance~\cite{kang2019incentive, xu2021besifl, gao2022fgfl, 9997114}.

Despite these advances, CrowdFL systems remain highly vulnerable to adversarial behavior. In particular, stealthy attackers~\cite{ModelReplacement, A3FL, shejwalkar2021manipulating, IBA, neurotoxin} can craft malicious updates that closely resemble benign ones, allowing them to evade standard detection mechanisms such as robust aggregation~\cite{blanchard2017machine} or anomaly filtering~\cite{Flame}. Over time, such adversaries can gradually accumulate reputation and gain influence in future training rounds, ultimately degrading model performance or embedding targeted backdoors. Existing works primarily focus on improving detection~\cite{Flame, rieger2022deepsight, huang2023multi} or incentive design\cite{dif2025autodfl,kang2019incentive,xu2021besifl,gao2022fgfl,9997114,feng2022crowdfl}, but largely overlook the fundamental question of how to quantify and control the robustness of reputation mechanisms under adaptive and evasive adversaries. Furthermore, recent reputation models (\eg, AutoDFL~\cite{dif2025autodfl}, SSMTD~\cite{li2026robust}) are often decoupled from the actual robustness of the aggregation process. As a result, they may assign high trust scores to participants whose malicious behavior remains undetected, leading to a mismatch between perceived and actual trustworthiness. This gap becomes critical in crowdsourcing environments, where reputation directly influences participant selection and long-term system security and stability.

\smallskip
To address these limitations, we propose \textbf{R2CFL}, a robust reputation-driven CrowdFL framework. R2CFL introduces a novel reputation model that is tightly coupled with a detect-and-filter aggregation strategy, namely robust nearest neighbor mixing (R2-NNM). The key idea is to link reputation evolution to the verified statistical behavior of participant updates, ensuring that trust accumulation reflects the true robustness of the underlying defense. By doing so, R2CFL prevents stealthy adversaries from gradually increasing their influence while preserving the effectiveness of filtering mechanism. Extensive experiments on a proof-of-concept demonstrate that R2CFL achieves strong robustness against adaptive Byzantine and backdoor attacks. The proposed R2-NNM mechanism matches or outperforms existing defenses, while the reputation model accurately captures their true positive and false positive rates (TPR/FPR), providing a reliable measure of participant trustworthiness in dynamic CrowdFL environments.


\section{Related Work}
\label{sec:relatedwork}

Existing CrowdFL frameworks aim to improve trust, incentives, and scalability, yet they provide limited guarantees against adversarial behavior. A substantial body of work integrates reputation and incentive mechanisms to encourage honest participation. Smart contract-based frameworks such as FedCFB~\cite{chen2024credible} assess client contributions to mitigate poisoning and free-riding, while TWFL~\cite{yuan2024trustworthy} leverages homomorphic encryption to secure gradient exchange. Incentive-driven systems, including BeSFL~\cite{xu2021besifl}, FGFL~\cite{gao2022fgfl} adopts accuracy-based rewards, auditing mechanisms, and reputation-aware participant selection. PoIS~\cite{9997114} uses feature attribution techniques to identify high-quality contributors. Similarly, SSMTD~\cite{li2026robust} relies on decentralized oracles to compute reputation scores, but primarily focuses on consistency rather than adversarial robustness. More recently, DARTIC~\cite{bouchihadartic} introduced a privacy-preserving reputation framework for decentralized crowdsourcing. In its CrowdFL instantiation, worker contributions are assessed using HDBSCAN clustering, following the same principle as FLAME~\cite{Flame}, to identify anomalous updates.

\smallskip
While these approaches improve participant selection and discourage dishonest behavior, they largely rely on heuristic assessments of client quality rather than providing verifiable guarantees on the correctness of the submitted updates. Consequently, aggregation and validation remain dependent on probabilistic or subjective evaluation models. A complementary line of research investigates cryptographic mechanisms to make the FL process itself verifiable, rather than relying solely on reputation or economic incentives. SettleFL~\cite{liang2026settlefl} introduces an efficient reward settlement mechanism through two variants: optimistic execution (\ie, \textit{commit-and-challenge}) and SNARK-based validity proofs (\ie, \textit{commit-with-proof}). However, it still assumes that client contributions have been evaluated correctly. VerifBFL~\cite{bellachia2025verifbfl} extends this direction by introducing a recursive zero-knowledge proof framework for both training correctness (\ie, \textit{proof of accuracy}) and aggregation integrity. Nevertheless, its proof of accuracy remains vulnerable to stealthy local data manipulation, such as backdoor injection.

\smallskip
\textbf{Gap.} Despite these advances, existing approaches rely on coarse-grained or static contribution metrics and loosely integrate reputation with robust statistical validation. Consequently, they remain vulnerable to adaptive, stealthy attacks that evade detection while gradually accumulating trust. To the best of our knowledge, no existing CrowdFL framework explicitly quantifies the robustness or fairness of its reputation mechanism, nor evaluates whether reputation accurately reflects the effectiveness of the underlying defense under stealthy conditions.

\section{Proposed R2CFL}
\label{sec:method}

In this section, we present \textbf{R2CFL}, a robust reputation-driven CrowdFL framework designed to operate in the presence of adversarial workers. R2CFL addresses two fundamental limitations of existing CrowdFL systems: (i) vulnerability to adaptive and stealthy model manipulation and (ii) insufficient automation of evaluation and incentive management. 

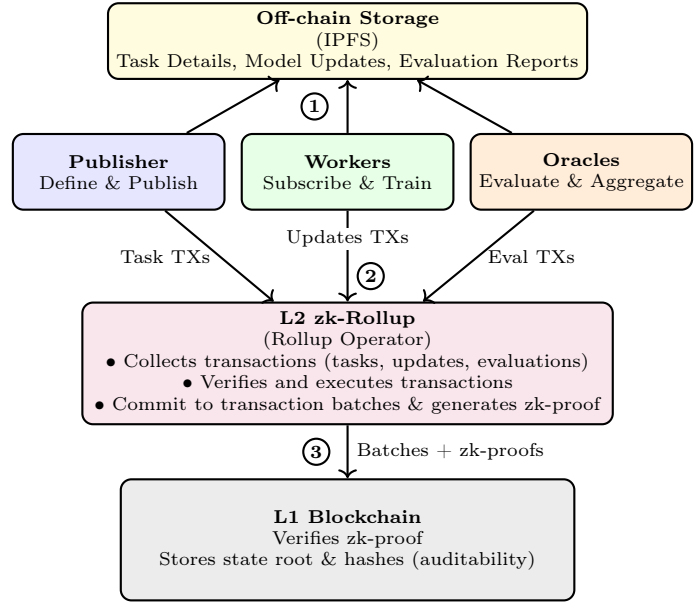
\begin{figure}[th]
\centering
\begin{tikzpicture}[
    node distance=1.2cm and 2.2cm,
    every node/.style={font=\scriptsize},
    box/.style={draw, rounded corners, thick, align=center, minimum width=2.8cm, minimum height=1cm},
    arrow/.style={->, thick},
    num/.style={circle, draw, fill=white, thick, inner sep=1.2pt, font=\scriptsize\bfseries}
]

\node[box, fill=blue!10] (publisher) {\textbf{Publisher} \\ Define \& Publish};
\node[box, fill=green!10, right=0.2cm of publisher] (workers) {\textbf{Workers} \\ Subscribe \& Train};
\node[box, fill=orange!15, right=0.2cm of workers] (oracles) {\textbf{Oracles} \\ Evaluate \& Aggregate};

\node[box, fill=yellow!15, above=0.7cm of workers] (storage)
{\textbf{Off-chain Storage} \\
\scriptsize (IPFS) \\ Task Details, Model Updates, Evaluation Reports};

\node[box, fill=purple!10, below=of workers, minimum width=6.5cm, minimum height=1.6cm] (l2) 
{\textbf{L2 zk-Rollup} \\ 
(Rollup Operator) \\
\scriptsize
\begin{tabular}{c}
$\bullet$ Collects transactions (tasks, updates, evaluations) \\
$\bullet$ Verifies and executes transactions \\
$\bullet$ Commit to transaction batches \& generates zk-proof
\end{tabular}
};

\node[box, fill=gray!15, below=0.7cm of l2, minimum width=6cm, minimum height=1.6cm] (l1)
{\textbf{L1 Blockchain} \\
\scriptsize
Verifies zk-proof \\
Stores state root \& hashes (auditability)
};

\draw[arrow] (publisher) -- node[left]{\scriptsize Task TXs} (l2);
\draw[arrow] (workers) -- 
    node[above, yshift=2pt, fill=white, inner sep=1pt]{\scriptsize Updates TXs}
    node[num, pos=0.72, right=0.12cm] {2}
    (l2);
\draw[arrow] (oracles) -- node[right]{\scriptsize Eval TXs} (l2);

\draw[arrow] (l2) -- node[right]{\scriptsize Batches + zk-proofs} (l1) node[num, pos=0.5, right=-0.6cm] {3};

\draw[arrow] (publisher) -- (storage);

\draw[arrow] (workers) -- (storage) node[num, pos=0.5, left=0.25cm] {1};

\draw[arrow] (oracles) -- (storage);

\end{tikzpicture}

\caption{Overview of an L2-scalable deployment of R2CFL.
\textbf{(1)} \textit{Task publishers}, \textit{workers}, and \textit{oracles} store large data objects, such as task details and model updates, off-chain via IPFS, whose content is referenced by cryptographic content hashes.
\textbf{(2)} They submit the corresponding smart contract TXs to an L2 zk-rollup, where the rollup operator processes and executes them.
\textbf{(3)} The L2 batches these TXs and generates a zk-proof attesting to the correctness of the resulting state transition, which is verified by the L1 blockchain. The L1 stores the corresponding state root for auditability and finalizes the committed state.}

\label{fig:r2cfl_overview}
\end{figure}

\subsection{System Model}

R2CFL system follows a generic crowdsourced FL paradigm that can operate in both \emph{centralized} and \emph{decentralized} settings. In the \textit{centralized setting}, a parameter server orchestrates tasks, coordinates training, performs evaluation, and aggregates model updates. In the \textit{decentralized setting}, R2CFL can be instantiated as a blockchained FL system (BFL)~\cite{kim2019blockchained} or a decentralized application (DApp)~\cite{bellachia2025verifbfl,dif2025autodfl}, where coordination is enforced through smart contracts, and evaluation and aggregation are delegated to a decentralized oracle network. To improve scalability, the blockchain deployment adopts a Layer-2 (L2) zk-rollup architecture, in which transactions are executed and batched off-chain before cryptographic proofs are submitted to the underlying Layer-1 (L1) blockchain for verification and finalization, as illustrated in Figure~\ref{fig:r2cfl_overview}. In this case, the blockchain and oracle layers are assumed to be secure and verifiable. The system involves four main actors:

\begin{itemize}
    \item \textbf{Publishers} (Requesters) define FL tasks, including the model architecture, training objectives, and reward policies.
    
    \item \textbf{Workers} (Clients) locally train models on private data and submit updates to the system.
    
    \item \textbf{Evaluators} (Aggregators) perform two core functions: 
    (i) \emph{evaluation}, where submitted updates are assessed using robust detect-and-filter mechanisms, and 
    (ii) \emph{aggregation}, where validated updates are combined into a global model using robust aggregation rules. 
    These roles are executed by the central server in \textit{centralized} settings, or by decentralized oracles in \textit{decentralized} settings.
    
    \item \textbf{Validators} (only in the decentralized setting) maintain the blockchain, verify transactions, and execute smart contract functions under a secure consensus protocol.
\end{itemize}


\subsection{Threat Model}

\textbf{Adversarial assumptions.} 
R2CFL operates in an open and potentially adversarial environment. Participants—including task publishers and workers—may behave strategically, deviate from the protocol, or collude to maximize utility.

In the \textit{centralized setting}, the server is assumed to be \emph{honest-but-curious}~\cite{xu2019verifynet}.  
In the \textit{decentralized setting}, the underlying blockchain and oracle protocols are assumed to be secure~\cite{dif2025autodfl}, while application-layer participants (\ie, \textit{task publishers} and \textit{workers}) remain untrusted. Under these assumptions, R2CFL considers the following threats:

\begin{itemize}

\item \textbf{Poisoning.} 
Adversarial workers may manipulate their local training data or updates to degrade model performance (\emph{untargeted attacks}) or embed hidden malicious behaviors (\emph{targeted/backdoor attacks}). These attacks can be adaptive and stealthy~\cite{A3FL, neurotoxin, shejwalkar2021manipulating}.

\item \textbf{Free-riding.} 
Workers may submit arbitrary, stale, or low-effort updates, or skip training while still attempting to receive rewards, thereby degrading convergence and fairness.

\end{itemize}


R2CFL mitigates these threats through robust reputation-driven detect-and-filter aggregation and verifiable coordination mechanisms. These mechanisms aim to ensure robustness, accountability, and fairness across both deployments. Since model evaluation and aggregation are performed either by an honest server or by a secure oracle network with verifiable execution, collusion between workers and task publishers is inherently mitigated. However, collusion involving the aggregation entities could undermine the integrity of the evaluation and aggregation procedures. Such scenarios require compromising the underlying infrastructure and are therefore considered outside the scope of this work.

\subsection{R2CFL Workflow}
\label{sec:architecture}

The R2CFL system depicted in Figure~\ref{fig:r2cfl_overview} follows the CrowdFL design presented in AutoDFL~\cite{dif2025autodfl} and operates through six global phases. \textbf{First}, a task publisher defines the learning objective, model, validation data, reward policy, and constraints, then publishes metadata on-chain while storing the full content on an off-chain storage system (\eg, IPFS) and locking rewards in a smart contract escrow. \textbf{Second}, workers subscribe to the task and are selected based on on-chain reputation and eligibility, potentially requiring collateral to deter malicious behavior. \textbf{Third}, selected workers perform local training on private data, optionally using differential privacy, and upload model updates to IPFS while only storing hashes on-chain to reduce blockchain overhead. \textbf{Fourth}, an oracle network retrieves updates and applies a robust filtering and aggregation mechanism to detect and exclude low-quality, free-riding, or poisoned contributions, with results anchored on-chain. \textbf{Fifth}, the aggregated global model is stored on IPFS and its hash is recorded on-chain for integrity, enabling workers to access it for subsequent rounds. \textbf{Finally}, after task completion, the oracle network updates worker reputations on-chain based on validated contributions, tightly coupling reputation evolution with the robustness of the filtering process.

\subsection{R2-NNM Filtering}
\label{sec:reputationmodel_robust}

Classical reputation mechanisms in federated learning rely on performance signals (\eg, validation accuracy~\cite{dif2025autodfl}), probabilistic interaction outcomes (\eg, subjective logic and Shapley values~\cite{an2024freb}), or trust inference (\eg, based on Markov processes and truth discovery~\cite{li2026robust}). While convenient, these mechanisms become unreliable in stealthy adversarial settings, where the underlying signals can be manipulated through sophisticated attacks such as backdoor attacks~\cite{A3FL, neurotoxin} and strategic model poisoning~\cite{shejwalkar2021manipulating}. To address these limitations, we propose a \emph{reputation-driven} robust geometric filtering mechanism, denoted as \textbf{R2-NNM} (Robust Reputation-Aware Nearest Neighbor Mixing). Instead of evaluating contributions through external metrics, filtering is inferred directly from the structural consistency of model updates within the population. 


At each FL round $t$, the aggregator collects a set of local updates
\[
\Theta^t = \{\theta_1^t, \ldots, \theta_n^t\}.
\]

Each update is first embedded into a vector space and processed through a two-stage robust pipeline:

\paragraph*{(1) Robust Reputation-Aware Nearest Neighbor Mixing.}
Each update is locally smoothed using its $k$ nearest neighbors:

\begin{equation}
\label{eq:mixing}
\tilde{\theta}_i^t =
\sum_{j \in \mathcal{N}_k(i) \cup \{i\}}
w_{ij} \theta_j^t,
\end{equation}

\noindent where $\mathcal{N}_k(i)$ denotes the set of nearest neighbors of client $i$, and the weights $w_{ij}$ are derived from current local reputation scores:

\begin{equation}
w_{ij}
=
(1 - \gamma_t)\cdot \frac{1}{|\mathcal{N}_k(i)\cup\{i\}|}
+
\gamma_t \cdot
\frac{L_{\mathrm{rep},j}^t}{\sum_{l \in \mathcal{N}_k(i)\cup\{i\}} L_{\mathrm{rep},l}^t},
\quad j \in \mathcal{N}_k(i)\cup\{i\}.
\end{equation}

This step acts as a \emph{geometric denoising operator} designed to reduce the influence of isolated or adversarial updates while preserving local consensus structures. The local reputation scores $L_{\text{rep}}$ are initialized uniformly to a neutral value (\eg, $0.5$) for all participants and are updated at the end of each round, with a progressive activation mechanism (as detailed in Sect.~\ref{sec:lcl}). The mixing weights are modulated by a time-dependent factor $\gamma_t$, defined as
\begin{equation}
\label{eq:warmup}
\gamma_t = \kappa \cdot \frac{t}{T_{\text{warm}}},
\end{equation}
which ensures a gradual transition from uniform to reputation-aware aggregation. Consequently, during the initial rounds, the influence of reputation is inactive, and the aggregation reduces to uniform weighting over the neighborhood:
\begin{equation}
w_{ij} = \frac{1}{|\mathcal{N}_k(i) \cup \{i\}|}.
\end{equation}

After the warm-up period (\ie, $t > T_{\text{warm}}$), the weights become increasingly influenced by the current $L_{\text{rep}}$ scores, yielding a reputation-aware mixing similar in spirit to NNM~\cite{allouah2023fixing}, but with a controlled and gradual activation of reputation information.

\paragraph*{(2) Robust Multi-Krum Selection.}
After mixing, a robust selection is applied:

\begin{equation}
\mathcal{C}_{\text{trusted}}^t = \text{MultiKrum}(\{\tilde{\theta}_i^t\}),
\end{equation}

which selects the subset of updates minimizing pairwise distances. This step provides guarantees against Byzantine behaviors under bounded adversarial assumptions~\cite{allouah2023fixing}.

Unlike classical defenses (\eg, FLAME or DeepSight), R2-NNM introduces a \emph{closed-loop interaction} between reputation and robustness: reputation influences filtering, and filtering in turn shapes reputation. Consequently, participants that are consistently rejected by the robust filtering stage experience a progressive loss of influence, while consistently accepted participants accumulate trust. This feedback mechanism makes it significantly more difficult for adversaries to progressively infiltrate the system through repeated unsuccessful attempts.

\subsection{Data-Free Reputation}
\label{sec:reputation_update}
We propose a \emph{data-free reputation model} that relies exclusively on the robust geometric filtering mechanism, \textbf{R2-NNM}. The key idea is: \emph{trust is not measured by performance, but by agreement with robust consensus under adversarial conditions}. This shift removes the need for unverifiable assumptions while strengthening robustness against adaptive adversaries.

\paragraph*{(1) Robust Local Reputation (Data-Free).} \label{sec:lcl}

We define reputation purely based on \emph{filter acceptance} and \emph{geometric consistency}. This eliminates any dependency on external validation datasets~\cite{dif2025autodfl}. Each client is assigned a binary indicator:
\begin{equation}
\text{Reliable}_i^t =
\begin{cases}
1 & \text{if } c_i \in \mathcal{C}_{\text{trusted}}^t \\
0 & \text{otherwise}
\end{cases}
\end{equation}
where $\mathcal{C}_{\text{trusted}}^t$ denotes the set of accepted clients at round $t$. To refine the binary selection signal induced by acceptance rule, we define a soft reliability score. Instead of relying on raw update distances~\cite{dif2025autodfl} or cosine similarity to a reference model~\cite{li2026robust}, we leverage pairwise distances between client updates and compute a robust normalized deviation based on the median and median absolute deviation (MAD):

\begin{equation}
\text{scaled}_i^t =
\frac{s_i^t - \mathrm{median}(s^t)}{\mathrm{MAD}(s^t) + \epsilon}.
\end{equation}

We then define a soft trust signal using a sigmoid transformation:
\begin{equation}
\hat{C}_i^t =
\sigma\!\left(-\frac{\text{scaled}_i^t}{\tau}\right),
\end{equation}
where $\tau$ is a temperature parameter controlling the sensitivity of the transformation, and $\sigma(\cdot)$ denotes the sigmoid function. A lower value of $\text{scaled}_i^t$ therefore yields a higher $\hat{C}_i^t$, assigning greater soft trust to clients whose scores are closer to or below the median. To prevent over-penalization of rejected clients while preserving separation between selected and non-selected participants, we combine both signals as:
\begin{equation}
\label{eq:contrib}
C_i^t = \alpha \cdot \hat{C}_i^t + (1 - \alpha) \cdot \text{Reliable}_i^t , 
\end{equation}

where $\alpha \in [0,1]$ controls the relative influence of $\hat{C}_i^t$ and $\text{Reliable}_i^t$. The robust local reputation is then updated using an exponential moving average (EMA) with warm-up stabilization:

\begin{equation}
\label{eq:robust_lcl_rep}
L_{\mathrm{rep}, i}^t =
\beta_t L_{\mathrm{rep}, i}^{t-1}
+
(1 - \beta_t)\, C_i^t, \quad \beta_t = 0.95 \cdot (1-\gamma_t).
\end{equation}

We set the base momentum coefficient to $\beta_0=0.95$ to prioritize historical consistency over short-term fluctuations. This choice reduces variance amplification in stochastic and non-IID settings while preserving responsiveness to long-term behavioral drift. This formulation ensures that: \begin{inparaenum}[(i)]
\item updates consistently rejected by the robust filter contribute near \emph{zero} reputation.
\item consistent updates accumulate reputation proportionally to their agreement with the consensus.
\item no external supervision (\ie, data) is required.
\end{inparaenum} From a stochastic perspective, reputation becomes an empirical estimate of the probability of being consistently selected by a robust aggregation rule.

\paragraph*{(2) Task-Level Reputation Update.} Global reputation is updated at the end of each task based on the consistency of each participant's contribution with respect to the aggregated outcome. The update rule follows an \textit{asymmetric} EMA:

\begin{equation}
R_i^{t+1} =
\begin{cases}
\omega_i \psi \, R_i^t + (1 - \omega_i \psi)\, L_{\text{rep},i}, 
& \text{if } L_{\text{rep}, i} \ge R_{\min}, \\[6pt]
\omega_i \xi \, R_i^t + (1 - \omega_i \xi)\, L_{\text{rep},i}, 
& \text{otherwise},
\end{cases}
\end{equation}

\begin{equation}
R_{\min} = \frac{1}{N} \sum_{j=1}^{N} R_j^t,
\end{equation}

where $L_{\text{rep}}$ denotes the reputation signal derived from robust aggregation, $R_{\min}$ is a minimum acceptance threshold, and $\psi > \xi$ control the asymmetric impact of positive and negative feedback~\cite{bouchiha2024llmchain}. The adaptive weighting factor $\omega_i \in (0,1)$ depends on the overall task participation:

\begin{equation}
\omega_i
=
\tanh\!\left(\frac{\lambda T_i}{2}\right),
\end{equation}

The participation factor $\omega_i$ increases monotonically with the number of completed tasks $T_i$ with $\lambda$ controlling the scale of sensitivity to participation. 

The mechanism creates a reputation process that is \textit{adaptive}, \textit{participation-aware}, and \textit{asymmetric}, allowing trust to be accumulated gradually while enabling rapid correction when participants deviate from the collective consensus. 

\section{Experimental Evaluation}
\label{sec:experiments}

\quad To evaluate the R2CFL framework, we developed a proof of concept. The code with all the technical details are available on GitHub\footnote{\href{https://github.com/mohaminemed/R2CFL}{https://github.com/mohaminemed/R2CFL}}. All experiments are conducted on two identical NVIDIA L40S GPUs to ensure consistent conditions. 
\subsection{Experimental Setup}

\textit{Datasets and models.} We evaluate performance across three widely used benchmark datasets, namely \texttt{FashionMNIST}~\cite{xiao2017fashion}, \texttt{GTSRB}~\cite{gtsrb}, and \texttt{CIFAR-10}~\cite{cifar}, using adapted neural network architectures tailored to each task. For \texttt{FashionMNIST} (28$\times$28 grayscale images), we employ a lightweight convolutional neural network (Fashion\_CNN) consisting of two convolutional blocks followed by two fully connected layers. For \texttt{CIFAR-10} (32$\times$32 RGB images), we adopt a standard \texttt{ResNet18} architecture. For \texttt{GTSRB} (color traffic sign images), we use a CNN model optimized for the dataset’s characteristics. 

\textit{CrowdFL configuration.} Across all datasets, workers use an SGD optimizer, a learning rate of 0.01, momentum of 0.9, a batch size of 32, and one local epoch (\ie, \textit{vanilla FL}) per communication round. We run 200 communication rounds for \texttt{CIFAR-10}, and 100 rounds for \texttt{GTSRB} and \texttt{FashionMNIST}. Unless otherwise specified, the total number of workers is 20, 6 of which are malicious. We simulate data heterogeneity of workers by sampling from the original dataset using a Dirichlet distribution of parameter $\alpha_D=0.5$.

\textit{Attack configurations.} We evaluate several representative poisoning and backdoor attacks to ensure a comprehensive comparison, including optimized model poisoning (OMP)~\cite{shejwalkar2021manipulating}, Neurotoxin~\cite{neurotoxin}, and A3FL~\cite{A3FL}. OMP performs adversarial update manipulation by aligning the update with a worst-case direction under a stealth constraint. Neurotoxin suppresses gradients of high-importance parameters identified via gradient magnitude analysis, while A3FL jointly optimizes input-space triggers and performs poisoned training with label flipping on a subset of local data. A3FL and Neurotoxin are activated within a configurable window ($t\in[60-80]$, see Sect.~\ref{app:ablation} for a sensitivity analysis), whereas OMP remains active throughout the training process to achieve stronger poisoning.

\textit{Baselines.} We consider state-of-the-art reputation models (SSMTD~\cite{li2026robust}, AutoDFL~\cite{dif2025autodfl}), as well as robust FL defenses (M-Krum~\cite{blanchard2017machine}, FLAME~\cite{Flame}, DeepSight~\cite{rieger2022deepsight}, NNM~\cite{allouah2023fixing}) baselines to evaluate the effectiveness of our method. 

\textbf{SSMTD}~\cite{li2026robust} computes a client reputation score based on cosine similarity between updates and a global reference model, combined with a hidden Markov model (HMM) that capture temporal behavior. Specifically, each client update is evaluated using a convex combination of instantaneous similarity and an exponentially smoothed trust estimate. The resulting scores are aggregated across clients, and a robust decision rule based on the \textit{median} is used to filter updates.

 \textbf{AutoDFL}~\cite{dif2025autodfl} evaluates each client update by directly measuring its impact on model performance. The score is defined as the improvement in accuracy and reduction in loss relative to the baseline model. Clients are then ranked according to these utility scores, and the top-$m$ updates are selected for aggregation, where $k$ is defined as a fixed fraction of participating workers (for $n=20$, $m=13$). 

\textbf{DeepSight}~\cite{rieger2022deepsight} is a clustering-based defense that analyzes client updates through multiple complementary feature spaces, including NEUP (last-layer perturbation evidence), DDif (decision differences under synthetic noise), and cosine-directional similarity. Each feature space is independently clustered using HDBSCAN, and a consensus-based agreement across clustering views is used to identify \textit{low-trust} clusters as potentially malicious. Benign updates are further refined using \textit{median clipping} before aggregation. 

\textbf{FLAME}~\cite{Flame} is a clustering-based defense that operates on flattened last-layer model weights using cosine similarity and HDBSCAN to separate benign and malicious updates. The \textit{largest cluster} is assumed to correspond to benign clients, while smaller clusters are discarded as potential attacks. To improve robustness, FLAME applies \textit{norm clipping} followed by adaptive Gaussian \textit{noise injection} proportional to the median update norm. 

\textbf{Multi-Krum} (M-Krum)~\cite{blanchard2017machine} is a robust aggregation rule that selects updates based on pairwise Euclidean distances. For each client update, a score is computed as the sum of distances to its $(n - f - 2)$ nearest neighbors, where $f$ is the assumed number of Byzantine clients. The $m$ updates with the lowest scores are selected and aggregated using FedAvg.
For $n=20$, we set $m=13$ to minimize false positives, and we stress-test this choice by decreasing m, which increases filtering aggressiveness (see trade-off analysis in Sect.~\ref{app:ablation}).

\textbf{NNM}~\cite{allouah2023fixing} a nearest-neighbor mixing strategy, where each client update is refined by averaging it with its $k$ nearest neighbors in parameter space. This method acts as a local smoothing operator over model updates to reduce variance before applying a robust aggregation rule (\ie, M-Krum, for $n=20$, $m=13$).

\begin{figure*}
    \centering
    \subfloat[\texttt{FashionMNIST}]{
        \includegraphics[width=0.95\linewidth]{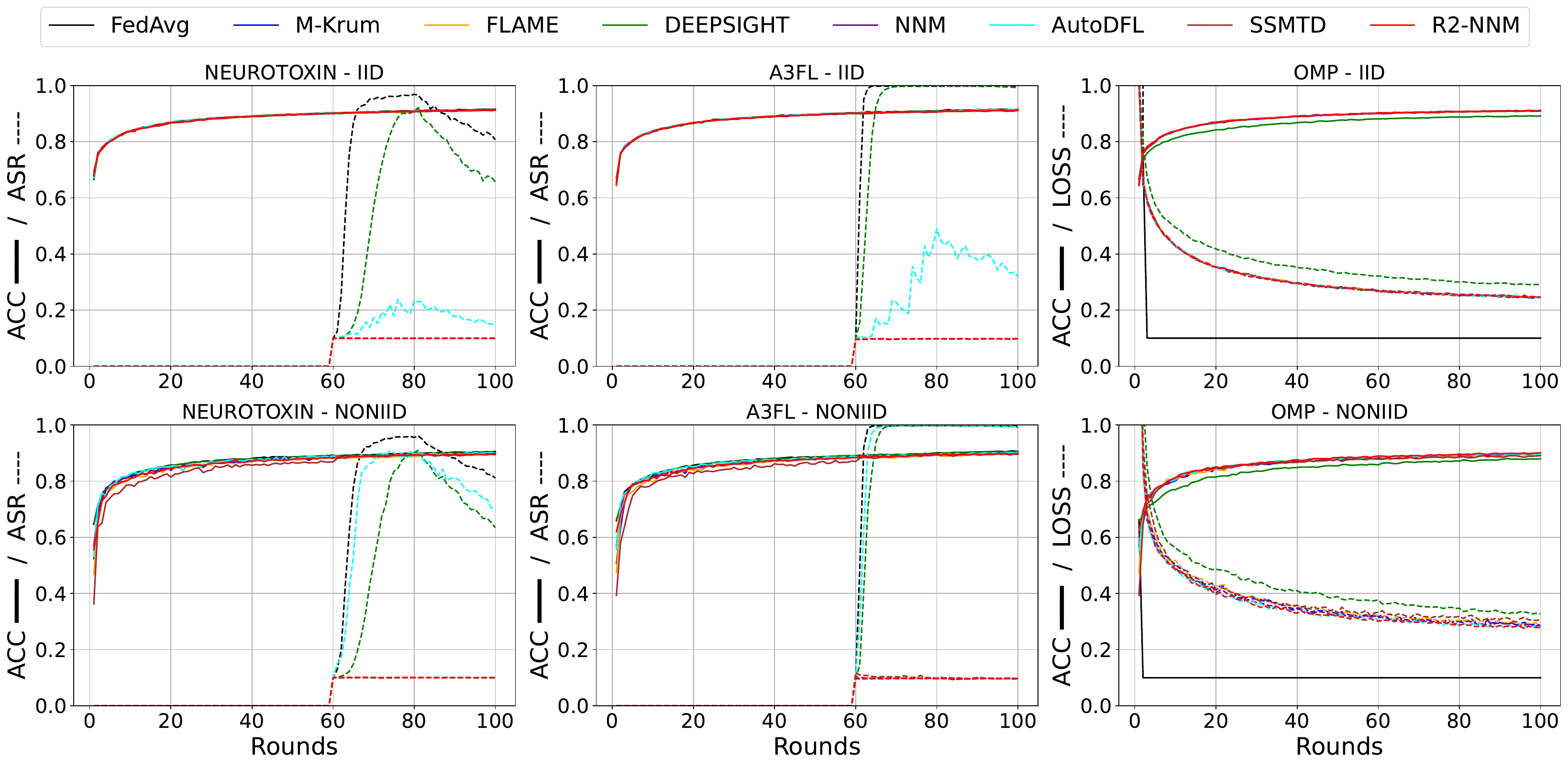}
        \label{fig:fashion_con}
}

\subfloat[\texttt{GTSRB}]{
        \includegraphics[width=0.95\linewidth]{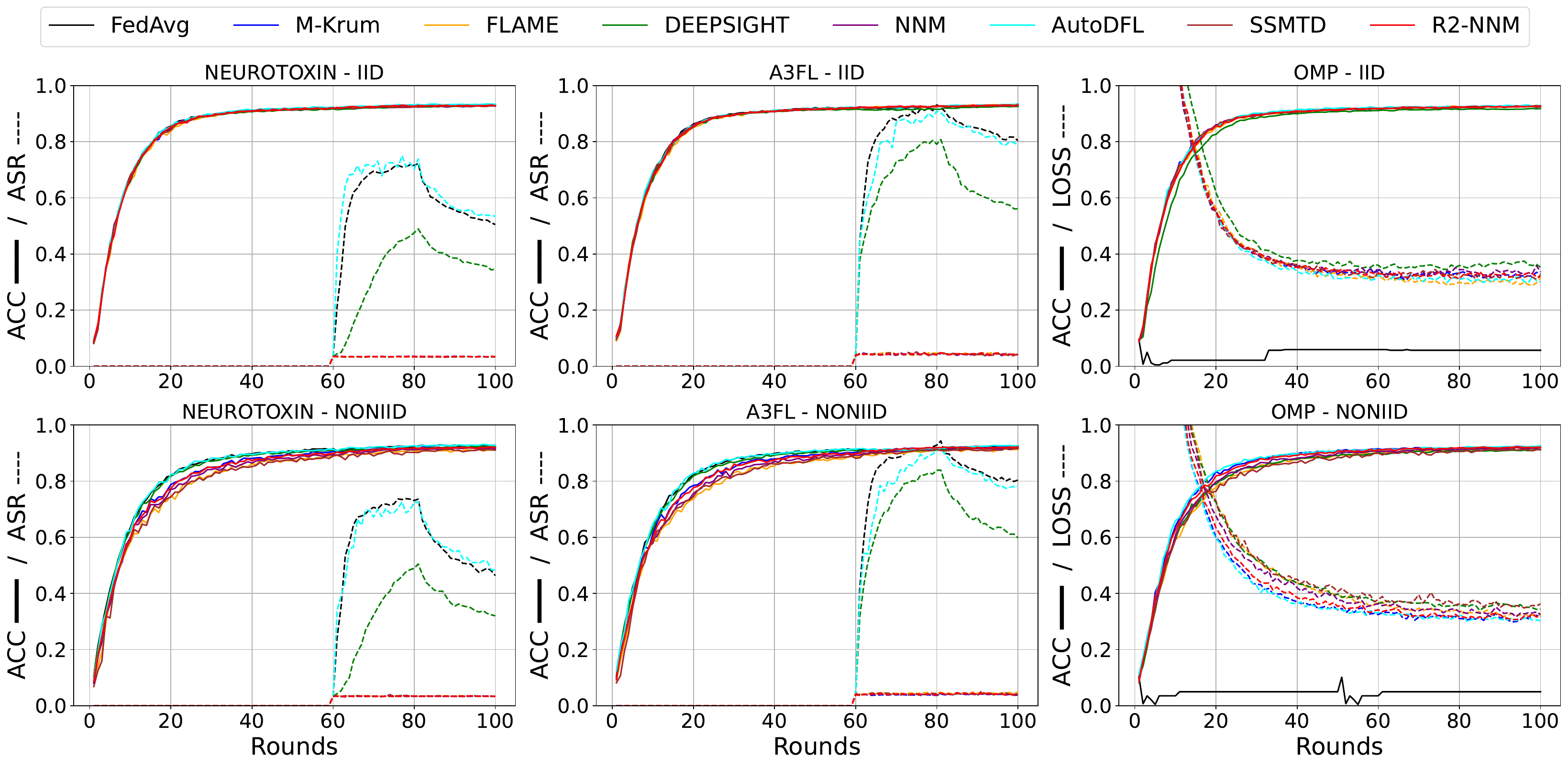}
        \label{fig:gtsrb_con}
}

\subfloat[\texttt{CIFAR-10}]{
        \includegraphics[width=0.95\linewidth]{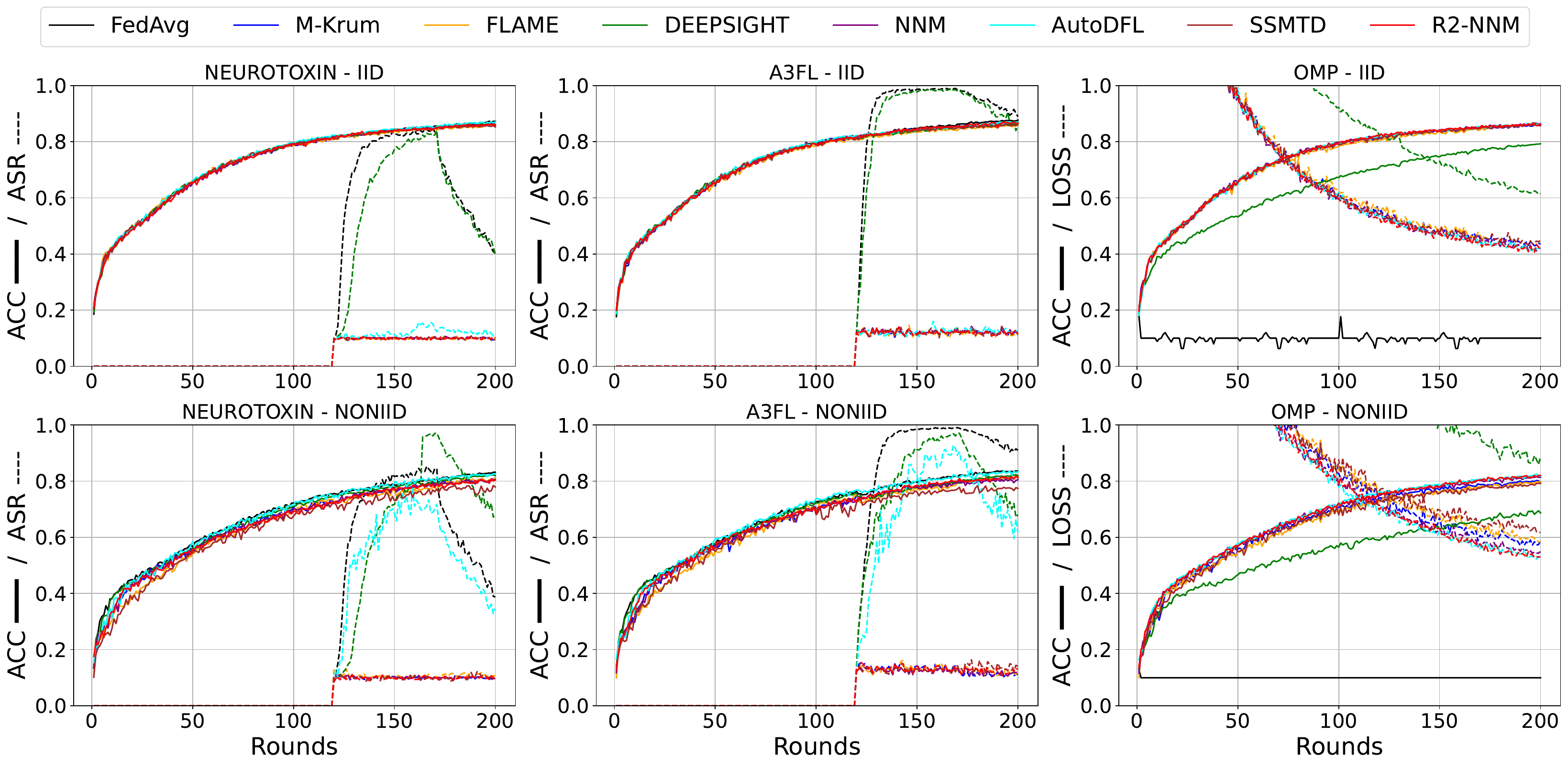}
        \label{fig:cifar_con}
}  
\caption{Convergence (ACC/LOSS) \& robustness (ASR) performance under attacks.}
\label{fig:robustness_perfo}
\end{figure*}

\begin{table}[t]
\centering
\caption{Detection (TPR/FPR) performance under attacks. \textbf{Bold} = best on both or FPR, \textcolor{brown}{brown} = 2nd best (TPR higher is better, FPR lower is better; DeepSight excluded from ranking as it detects nothing).}
\label{tab:tpr_fpr}
\setlength{\tabcolsep}{3pt}
\resizebox{0.95\textwidth}{!}{%
\begin{tabular}{l @{\hspace{8pt}} cc @{\hspace{6pt}} cc @{\hspace{16pt}} cc @{\hspace{6pt}} cc @{\hspace{16pt}} cc @{\hspace{6pt}} cc}
\toprule
& \multicolumn{4}{c}{Neurotoxin} & \multicolumn{4}{c}{A3FL} & \multicolumn{4}{c}{OMP} \\
\cmidrule(lr){2-5} \cmidrule(lr){6-9} \cmidrule(lr){10-13}
& \multicolumn{2}{c}{IID} & \multicolumn{2}{c}{Non-IID} & \multicolumn{2}{c}{IID} & \multicolumn{2}{c}{Non-IID} & \multicolumn{2}{c}{IID} & \multicolumn{2}{c}{Non-IID} \\
\cmidrule(lr){2-3} \cmidrule(lr){4-5} \cmidrule(lr){6-7} \cmidrule(lr){8-9} \cmidrule(lr){10-11} \cmidrule(lr){12-13}
Defense & TPR & FPR & TPR & FPR & TPR & FPR & TPR & FPR & TPR & FPR & TPR & FPR \\
\midrule
\rowcolor{blue!8} \multicolumn{13}{c}{\texttt{FashionMNIST}} \\
\midrule
M-Krum & 0.960 & 0.256 & 0.960 & 0.256 & \textcolor{brown}{0.963} & \textcolor{brown}{0.256} & 0.960 & 0.256 & 0.995 & 0.144 & 0.992 & 0.146 \\
FLAME & 0.976 & 0.410 & 0.976 & 0.413 & \textcolor{brown}{0.968} & \textcolor{brown}{0.410} & 0.963 & 0.414 & 0.995 & 0.206 & 0.995 & 0.216 \\
DeepSight & 0.000 & 0.000 & 0.000 & 0.000 & 0.000 & 0.000 & 0.000 & 0.000 & 0.000 & 0.000 & 0.000 & 0.000 \\
NNM & \textcolor{brown}{0.970} & \textcolor{brown}{0.255} & \textcolor{brown}{0.963} & \textcolor{brown}{0.255} & \textbf{0.968} & \textbf{0.255} & \textcolor{brown}{0.960} & \textcolor{brown}{0.256} & \textcolor{brown}{0.993} & \textcolor{brown}{0.003} & \textbf{0.993} & \textbf{0.003} \\
AutoDFL & 0.857 & 0.263 & 0.492 & 0.287 & 0.921 & 0.258 & 0.492 & 0.287 & \textcolor{brown}{0.993} & \textcolor{brown}{0.003} & \textbf{0.993} & \textbf{0.003} \\
SSMTD & 0.976 & 0.468 & 0.968 & 0.469 & 0.969 & 0.468 & 0.968 & 0.469 & 0.995 & 0.288 & 0.993 & 0.289 \\
\textbf{R2-NNM} & \textbf{0.976} & \textbf{0.255} & \textbf{0.968} & \textbf{0.255} & \textbf{0.968} & \textbf{0.255} & \textbf{0.962} & \textbf{0.255} & \textbf{0.995} & \textbf{0.002} & \textcolor{brown}{0.992} & \textcolor{brown}{0.004} \\
\midrule
\rowcolor{blue!8} \multicolumn{13}{c}{\texttt{GTSRB}} \\
\midrule
M-Krum & 0.960 & 0.256 & 0.960 & 0.256 & \textcolor{brown}{0.960} & \textcolor{brown}{0.256} & 0.960 & 0.255 & 0.995 & 0.073 & 0.993 & 0.003 \\
FLAME & \textcolor{brown}{0.984} & \textcolor{brown}{0.382} & 0.968 & 0.381 & 0.968 & 0.387 & 0.968 & 0.380 & 0.995 & 0.164 & 0.993 & 0.154 \\
DeepSight & 0.000 & 0.000 & 0.000 & 0.000 & 0.000 & 0.000 & 0.000 & 0.000 & 0.000 & 0.000 & 0.000 & 0.000 \\
NNM & \textcolor{brown}{0.960} & \textcolor{brown}{0.235} & \textcolor{brown}{0.960} & \textcolor{brown}{0.235} & \textcolor{brown}{0.960} & \textcolor{brown}{0.235} & \textcolor{brown}{0.952} & \textcolor{brown}{0.236} & \textbf{0.995} & \textbf{0.002} & \textbf{0.992} & \textbf{0.001} \\
AutoDFL & 0.183 & 0.308 & 0.389 & 0.294 & 0.492 & 0.287 & 0.532 & 0.284 & 0.852 & 0.064 & 0.940 & 0.026 \\
SSMTD & 0.984 & 0.467 & 0.968 & 0.469 & 0.976 & 0.468 & 0.968 & 0.469 & 0.997 & 0.287 & 0.997 & 0.287 \\
\textbf{R2-NNM} & \textbf{0.960} & \textbf{0.229} & \textbf{0.960} & \textbf{0.229} & \textbf{0.960} & \textbf{0.229} & \textbf{0.960} & \textbf{0.229} & \textbf{0.995} & \textbf{0.002} & \textcolor{brown}{0.990} & \textcolor{brown}{0.001} \\
\midrule
\rowcolor{blue!8} \multicolumn{13}{c}{\texttt{CIFAR-10}} \\
\midrule
M-Krum & 0.993 & 0.253 & \textcolor{brown}{0.990} & \textcolor{brown}{0.253} & 0.987 & 0.253 & \textcolor{brown}{0.984} & \textcolor{brown}{0.253} & \textcolor{brown}{0.998} & \textcolor{brown}{0.144} & 0.998 & 0.144 \\
FLAME & 0.987 & 0.400 & 0.990 & 0.403 & 0.990 & 0.399 & 0.987 & 0.403 & 0.998 & 0.214 & 0.998 & 0.214 \\
DeepSight & 0.000 & 0.000 & 0.000 & 0.000 & 0.000 & 0.000 & 0.000 & 0.000 & 0.000 & 0.000 & 0.000 & 0.000 \\
NNM & \textbf{0.993} & \textbf{0.243} & \textcolor{brown}{0.987} & \textcolor{brown}{0.243} & \textcolor{brown}{0.990} & \textcolor{brown}{0.243} & \textbf{0.984} & \textbf{0.243} & \textbf{0.998} & \textbf{0.001} & \textcolor{brown}{0.997} & \textcolor{brown}{0.001} \\
AutoDFL & 0.931 & 0.248 & 0.542 & 0.280 & 0.977 & 0.244 & 0.533 & 0.281 & \textbf{0.998} & \textbf{0.001} & 0.995 & 0.002 \\
SSMTD & 0.990 & 0.459 & 0.987 & 0.460 & 0.984 & 0.460 & 0.987 & 0.460 & 0.998 & 0.286 & 0.998 & 0.286 \\
\textbf{R2-NNM} & \textbf{0.993} & \textbf{0.243} & \textbf{0.990} & \textbf{0.243} & \textbf{0.989} & \textbf{0.222} & \textbf{0.984} & \textbf{0.243} & \textbf{0.998} & \textbf{0.001} & \textbf{0.998} & \textbf{0.001} \\
\bottomrule
\end{tabular}%
}
\end{table}
\subsection{Convergence \& Detection Performance}

\paragraph{Metrics.} We evaluate the convergence \& detection performance using:
\begin{itemize}
\item \textit{Main-task metrics (ACC/LOSS):} We report the average main task accuracy (ACC) and loss (LOSS) on clean test data~\cite{A3FL}.
\item \textit{Attack success rate (ASR):} We report the ASR dynamics, defined as the proportion of poisoned test samples that are misclassified into the target label~\cite{A3FL}.
\item \textit{True Positive Rate (TPR):} The proportion of malicious clients correctly identified by the detection mechanism.
\item \textit{False Positive Rate (FPR):} The proportion of benign clients incorrectly classified as malicious by the detection mechanism.
\end{itemize}
The results in Figure~\ref{fig:robustness_perfo} and Table~\ref{tab:tpr_fpr} are reported across multiple attack scenarios (Neurotoxin~\cite{neurotoxin}, A3FL~\cite{A3FL}, and OMP~\cite{shejwalkar2021manipulating}) and data distributions (IID and non-IID). An ideal defense achieves a high TPR while maintaining a low FPR.

NNM and the proposed R2-NNM, consistently achieve the best trade-off between TPR and FPR across all settings. In particular, they maintain high detection rates (TPR $\approx$ 0.95--0.99) while keeping false positives relatively low. Notably, R2-NNM closely matches the performance of NNM in all scenarios, demonstrating that integrating reputation does not degrade detection capability while enabling trust-aware aggregation. M-Krum achieves comparable performance (ASR $<$ 0.10) and near-perfect detection in simple settings like OMP (TPR $\approx$ 0.99), with slight degradation under more challenging cases like A3FL on \texttt{GTSRB} (FPR $\approx$ 0.26). FLAME attains high TPR (up to 0.99) but at the cost of a high FPR (up to 0.38) due to aggressive filtering that impacts benign clients. In contrast, AutoDFL is unstable across attacks, with good performance under OMP but poor detection under backdoor attacks (TPR down to 0.18 under Neurotoxin and A3FL with ASR $\approx$ 1.00). SSMTD convergence is impacted under Non-IID due to strong median filtering. Finally, DeepSight consistently yields $(\mathrm{TPR}=0, \mathrm{FPR}=0)$, indicating failure to detect malicious clients.

The transition from IID to non-IID data has minimal impact on NNM-based methods. Both NNM and R2-NNM maintain nearly identical TPR and FPR values across distributions. This demonstrates strong robustness to data heterogeneity, which is critical in realistic scenarios. Note that the differences in the convergence under A3FL and Neurotoxin are negligible since the attacks does not impact the main task. Additionally, the impact of strong filtering on convergence (as observed with the median in SSMTD) is evident only under the Non-IID setting. This indicates that even when robustness and convergence are preserved—as reported in most studies—fairness toward discarded clients can be significantly affected.

\subsection{Reputation Effectiveness and Fairness}


For each trainer $i$, we simulate the robust local evaluation (Eq.~\eqref{eq:robust_lcl_rep}) over multiple FL rounds. The stochastic reliability indicator $\text{Reliable}_i^t$ is sampled using the defense-specific TPR and FPR:
\begin{multline}
\small 
\mathbb{P}[\text{Reliable}_i^t = 1 \mid i \in \mathcal{B}] = 1 - \text{FPR}, 
\quad
\mathbb{P}[\text{Reliable}_i^t = 1 \mid i \in \mathcal{M}] = 1 - \text{TPR}.
\end{multline}

This models the probability that a benign or malicious trainer is considered trustworthy by the defense.  

\begin{table*}[t]
\centering
\caption{Final $L_{\text{rep},i}^{t}$ scores at round 100 under A3FL (the \textit{worst-case} scenario). Best $\Delta$ is highlighted in \textbf{bold}, and second-best $\Delta$ is highlighted in \textcolor{brown}{brown}.}
\label{tab:rep_worstcase}
\resizebox{0.9\textwidth}{!}{
\begin{tabular}{lcccccc}
\hline
& \multicolumn{3}{c|}{\textbf{IID}} & \multicolumn{3}{c}{\textbf{Non-IID}} \\
\textbf{Method}
& $\bar{R}_{\mathcal{B}}$ & $\bar{R}_{\mathcal{M}}$ & \textbf{$\Delta$}
& $\bar{R}_{\mathcal{B}}$ & $\bar{R}_{\mathcal{M}}$ & \textbf{$\Delta$} \\

\hline
\rowcolor{blue!8} &  & & \texttt{FashionMNIST} & & & \\
\hline

M-Krum
& $0.797\pm0.049$ & $0.079\pm0.016$ & 0.718
& $0.796\pm0.048$ & $0.092\pm0.023$ & 0.704 \\

FLAME
& $0.688\pm0.058$ & $0.087\pm0.020$ & 0.601
& $0.684\pm0.055$ & $0.081\pm0.019$ & 0.603 \\

DeepSight
& $0.973\pm0.000$ & $0.759\pm0.003$ & 0.214
& $0.973\pm0.000$ & $0.760\pm0.002$ & 0.214 \\

NNM
& $0.813\pm0.048$ & $0.091\pm0.021$ & \textcolor{brown}{0.721}
& $0.805\pm0.051$ & $0.089\pm0.020$ & \textcolor{brown}{0.716} \\

AutoDFL
& $0.792\pm0.050$ & $0.120\pm0.034$ & 0.672
& $0.773\pm0.050$ & $0.419\pm0.057$ & 0.354 \\

SSMTD
& $0.644\pm0.056$ & $0.079\pm0.016$ & 0.565
& $0.650\pm0.055$ & $0.085\pm0.016$ & 0.566 \\

\textbf{R2-NNM}
& $0.822\pm0.048$ & $0.085\pm0.019$ & \textbf{0.737}
& $0.818\pm0.047$ & $0.096\pm0.025$ & \textbf{0.722} \\

\hline
\rowcolor{blue!8} &  & & \texttt{GTSRB} & & & \\
\hline

M-Krum
& $0.794\pm0.050$ & $0.084\pm0.019$ & 0.711
& $0.796\pm0.049$ & $0.085\pm0.020$ & 0.711 \\
FLAME
& $0.708\pm0.054$ & $0.085\pm0.019$ & 0.623
& $0.708\pm0.054$ & $0.084\pm0.018$ & 0.624 \\
DeepSight
& $0.973\pm0.001$ & $0.760\pm0.002$ & 0.213
& $0.973\pm0.000$ & $0.759\pm0.003$ & 0.214 \\
NNM
& $0.795\pm0.048$ & $0.083\pm0.022$ & \textbf{0.712}
& $0.809\pm0.049$ & $0.096\pm0.025$ & \textcolor{brown}{0.713} \\
AutoDFL
& $0.773\pm0.051$ & $0.414\pm0.061$ & 0.359
& $0.774\pm0.049$ & $0.392\pm0.055$ & 0.382 \\
SSMTD
& $0.647\pm0.055$ & $0.082\pm0.019$ & 0.565
& $0.647\pm0.059$ & $0.085\pm0.020$ & 0.562 \\
\textbf{R2-NNM}
& $0.794\pm0.050$ & $0.082\pm0.023$ & \textbf{0.712}
& $0.813\pm0.045$ & $0.090\pm0.023$ & \textbf{0.723} \\

\hline
\rowcolor{blue!8} &  & & \texttt{CIFAR-10} & & & \\
\hline

M-Krum
& $0.804\pm0.048$ & $0.074\pm0.014$ & 0.730
& $0.803\pm0.049$ & $0.076\pm0.014$ & 0.727 \\

FLAME
& $0.693\pm0.056$ & $0.071\pm0.012$ & 0.622
& $0.692\pm0.058$ & $0.072\pm0.013$ & 0.620 \\

DeepSight
& $0.973\pm0.000$ & $0.760\pm0.002$ & 0.214
& $0.973\pm0.000$ & $0.760\pm0.002$ & 0.214 \\

NNM
& $0.803\pm0.047$ & $0.071\pm0.012$ & \textcolor{brown}{0.733}
& $0.805\pm0.048$ & $0.076\pm0.015$ & \textcolor{brown}{0.730} \\

AutoDFL
& $0.804\pm0.049$ & $0.080\pm0.018$ & 0.724
& $0.778\pm0.052$ & $0.387\pm0.055$ & 0.390 \\

SSMTD
& $0.654\pm0.055$ & $0.075\pm0.014$ & 0.579
& $0.651\pm0.055$ & $0.072\pm0.011$ & 0.580 \\

\textbf{R2-NNM}
& $0.810\pm0.040$ & $0.076\pm0.015$ & \textbf{0.735}
& $0.806\pm0.048$ & $0.075\pm0.015$ & \textbf{0.731} \\

\hline

\end{tabular}}
\end{table*}

\paragraph*{Reputation Dynamics.}  
We then evaluate the local reputation by reporting these updates over sequential rounds, we track: \begin{inparaenum}[(i)]
    \item \textit{Malicious reputation decay:} For trainers $i \in \mathcal{M}$, the expected reputation decreases monotonically and converges toward zero.
    \item \textit{Benign reputation stability:} For trainers $i \in \mathcal{B}$, fluctuations are bounded by the FPR.
\end{inparaenum}

\paragraph*{Metrics.}  
We compute the final-round average reputation for each class:

\begin{equation}
\bar{R}_{\mathcal{B}} = \frac{1}{|\mathcal{B}|} \sum_{i \in \mathcal{B}} R_i^T,
\quad
\bar{R}_{\mathcal{M}} = \frac{1}{|\mathcal{M}|} \sum_{i \in \mathcal{M}} R_i^T,
\end{equation}

\noindent where $T$ denotes the final round (here $T=50$).  
By comparing $\bar{R}_{\mathcal{B}}$ and $\bar{R}_{\mathcal{M}}$ across defenses and data distributions, we directly measure:
\begin{itemize}
    \item \emph{Effectiveness:} The separation 
    $\Delta = \bar{R}_{\mathcal{B}} - \bar{R}_{\mathcal{M}}$ quantifies how strongly malicious trainers are penalized relative to benign ones.
    \item \emph{Fairness:} The stability of $\bar{R}_{\mathcal{B}}$ across IID and Non-IID settings, together with its variance, reflects whether honest trainers maintain high and stable reputation despite statistical heterogeneity.
\end{itemize}
The results in Table~\ref{tab:rep_worstcase} shows clear differences among defenses under stealthy attackers. M-Krum achieves strong and stable separation ($\Delta \approx 0.68$), with malicious reputations near zero. NNM and R2-NNM perform similarly or better ($\Delta \geq 0.70$), highlighting improved robustness and fairness through relaxed aggregation. FLAME maintains good separation ($\Delta \approx 0.61$) due to mild false positives, while SSMTD shows moderate separation ($\Delta \approx 0.57$) due to its less discriminative median-based filtering. AutoDFL degrades under non-IID, allowing high malicious reputation (up to $\approx 0.85$), reflecting poor separation under stealthy backdoor attacks. DeepSight fails entirely, with no separation ($\Delta \approx 0.21$). Overall, these trends align with TPR/FPR behavior: strong defenses ensure fast malicious decay and stable benign retention, whereas weaker ones reduce or lose discrimination, confirming the consistency of the reputation dynamics with theoretical expectations.

\subsection{Ablation Study} \label{app:ablation}
We study the impact of key system parameters, including the mixing parameter $k$ (Eq.\eqref{eq:mixing}), the total number of workers $n$, the number of malicious workers $f$, and the attack schedule. Appendices A–C further analyze time overhead, the effects of $\alpha$ (Eq.\eqref{eq:contrib}), and the warm-up phase (Eq.\eqref{eq:warmup}). 

\begin{figure}
    \centering
    \includegraphics[width=0.95\linewidth]{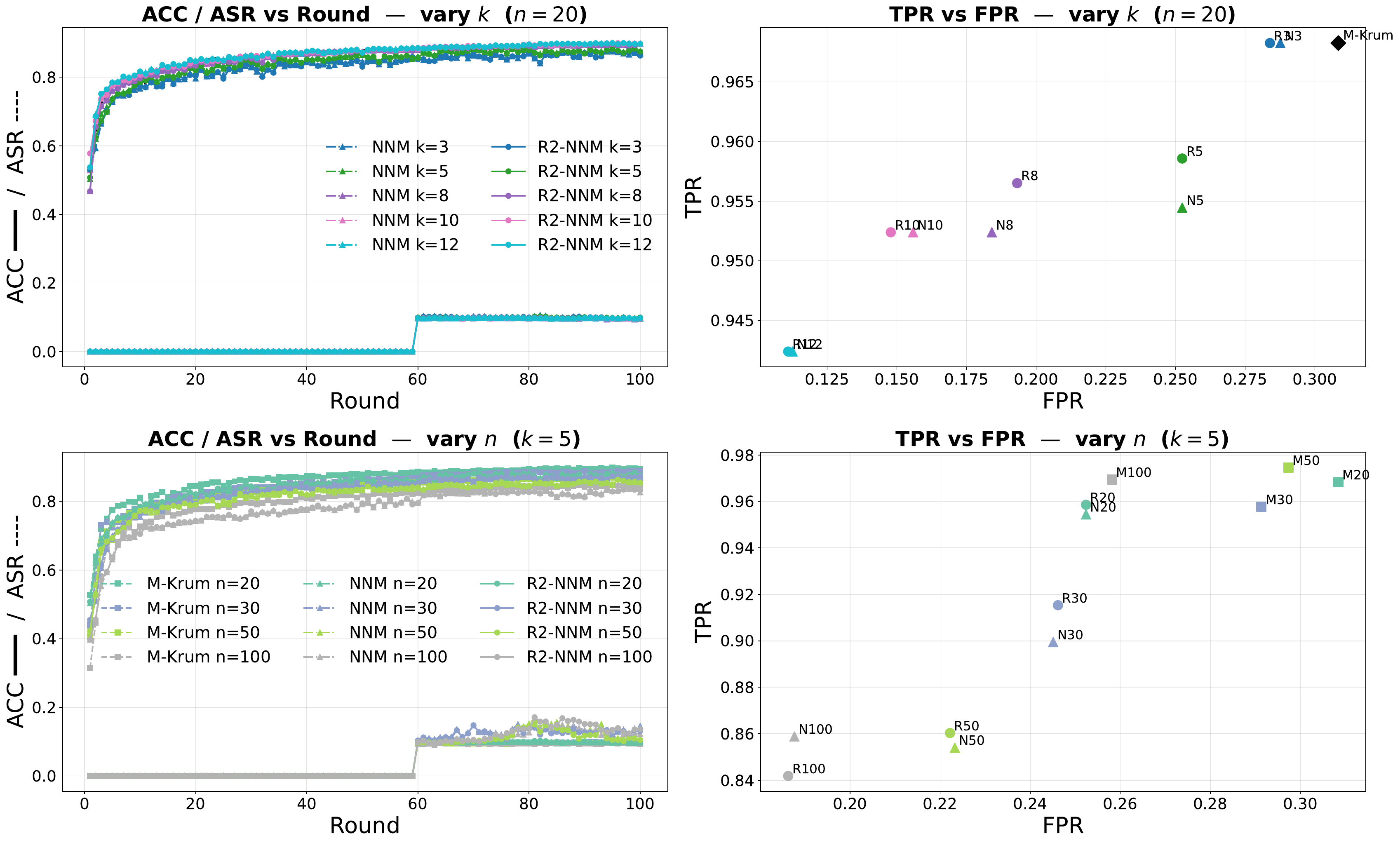}
    \caption{Impact of the mixing parameter $k$ (\textbf{top}) and the number of workers $n$ (\textbf{bottom}).}
    \label{fig:impact_k_n}
\end{figure}

\paragraph{Impact of the mixing parameter $k$.}
Figure~\ref{fig:impact_k_n} (top sub-figures) sweeps $k \in [3-12]$ for R2-NNM and NNM. As $k$ increases both methods exhibit a clear improvement in fairness, with FPR decreasing significantly (0.28 $\rightarrow$ 0.12). This regime reflects the benefit of stronger mixing, which reduces update variance, improves convergence stability, and enhances separation between benign and malicious contributions. Slight variations in TPR ($\Delta$TPR $\leq$ 0.02) are observed within the acceptable mixing range ($k\leq 10$), indicating that moderate mixing has only a limited impact on detection performance. Beyond this range, however, TPR degrades (down to 0.94) despite a further reduction in FPR ($<$ 0.12). This behavior suggests an over-smoothing effect, where excessive mixing increases conservativeness at the expense of detection sensitivity.

\paragraph{Impact of the total number of workers $n$.}
To evaluate scalability, we fix $k=5$ and increase the number of participants from $20$ to $100$. We see in Figure~\ref{fig:impact_k_n} (bottom sub-figures), as the federation grows, both NNM and R2-NNM experience a gradual decrease in accuracy (from $\approx87.5\%$ to $\approx83\%$) and a slight increase in ASR, reflecting the increased diversity of benign updates. Nevertheless, the FPR remains relatively stable ($0.22$--$0.26$), indicating that the detector maintains its fairness as the system scales. The main impact of increasing $n$ is on the TPR, which decreases from $0.97$ to $0.85$ for NNM and from $0.96$ to $0.86$ for R2-NNM. This behavior is expected, as larger federations reduce the statistical separability between benign and malicious updates. Importantly, the proposed reputation mechanism closely follows the underlying NNM detector without artificially improving or degrading its TPR/FPR trade-off.

\paragraph{Impact of the number of malicious clients $f$.}
Table~\ref{tab:f-ablation} sweeps the number of malicious clients
$f \in \{0, 3, 6\}$ for R2-NNM and M-Krum. At $f=0$ there is no attack,
so ASR is trivially 0 and TPR is trivially 1 for both defenses.

For $m=10$, R2-NNM maintains high robustness, with FPR decreasing from 0.26 to 0.23 as $f$ increases from 0 to 6. M-Krum shows a higher sensitivity to malicious presence in terms of false positives, with FPR remaining relatively high (0.50 $\rightarrow$ 0.47), although TPR remains stable around 0.97. This indicates that R2-NNM’s smoothed selection induces less conservative behavior, especially outside the attack window. For $m=13$, both methods benefit from the larger selection pool, resulting in overall lower FPR values. R2-NNM achieves a reduction from 0.19 to 0.16, while maintaining TPR around 0.96–1.0. Similarly, M-Krum improves its FPR from 0.35 to 0.30, while preserving competitive TPR values. Overall, reducing the selection count $m$ increases false
positives for both defenses, but the effect is roughly $1.5$--$2\times$
larger for M-Krum than for R2-NNM at every tested $f$, widening R2-NNM's
detection-precision advantage as fewer updates are selected per round.
Thus, R2-NNM's reputation mechanism is particularly valuable in
low-$m$ regimes, where purely distance-based selection has the least
information to work with.

\begin{table}[t]
\small
\begin{minipage}{0.44\textwidth}
\centering
\caption{Impact of $f$ (A3FL, \texttt{FashionMNIST}, non-IID).}
\label{tab:f-ablation}
\resizebox{0.9\columnwidth}{!}{
\begin{tabular}{l c c c c c c}
\toprule
\textbf{Defense} & \textbf{$m$} & \textbf{$f$} & \textbf{ACC} & \textbf{ASR} & \textbf{TPR} & \textbf{FPR} \\
\midrule
\multirow{6}{*}{M-Krum}
  & \multirow{3}{*}{13} & 0 & 0.894 & 0.000 & 1.000 & 0.350 \\
  &                     & 3 & 0.896 & 0.098 & 0.968 & 0.330 \\
  &                     & 6 & 0.897 & 0.096 & 0.962 & 0.308 \\
\cmidrule(lr){2-7}
  & \multirow{3}{*}{10} & 0 & 0.894 & 0.000 & 1.000 & 0.500 \\
  &                     & 3 & 0.894 & 0.097 & 0.968 & 0.485 \\
  &                     & 6 & 0.898 & 0.095 & 0.976 & 0.468 \\
  \midrule
  \multirow{6}{*}{\textbf{R2-NNM}}
  & \multirow{3}{*}{13} & 0 & 0.898 & 0.000 & 1.000 & \textbf{0.192} \\
  &                     & 3 & 0.898 & 0.097 & 0.968 & \textbf{0.178} \\
  &                     & 6 & 0.899 & 0.097 & 0.962 & \textbf{0.165} \\
\cmidrule(lr){2-7}
  & \multirow{3}{*}{10} & 0 & 0.898 & 0.000 & 1.000 & \textbf{0.269} \\
  &                     & 3 & 0.898 & 0.098 & 0.972 & \textbf{0.253} \\
  &                     & 6 & 0.899 & 0.097 & 0.972 & \textbf{0.234} \\
\bottomrule
\end{tabular}}
\end{minipage}
\begin{minipage}{0.48\textwidth}
\centering
\caption{Impact of attack-start schedule (A3FL, \texttt{FashionMNIST}, non-IID).}
\label{tab:attack-schedule-ablation}
\resizebox{0.95\columnwidth}{!}{
\begin{tabular}{l l c c c c}
\toprule
\textbf{Defense} & \textbf{Schedule} & \textbf{ACC} & \textbf{ASR} & \textbf{TPR} & \textbf{FPR} \\
\midrule
\multirow{3}{*}{M-Krum}
  & round 0  & 0.895 & 0.098 & 0.99 & 0.07 \\
  & round 30 & 0.897 & 0.097 & 0.99 & 0.18 \\
  & round 60 & 0.900 & 0.097 & 0.98 & 0.26 \\
\midrule
\multirow{3}{*}{NNM}
  & round 0  & 0.893 & \textcolor{brown}{0.098} & 0.99 & \textbf{0.00} \\
  & round 30 & 0.899 & \textcolor{brown}{0.097} & 0.99 & \textbf{0.07} \\
  & round 60 & 0.900 & \textbf{0.096} & 0.98 & \textbf{0.13} \\
\midrule
\multirow{3}{*}{\textbf{R2-NNM}}
  & round 0  & 0.898 & \textbf{0.097} & 0.99 & \textbf{0.00} \\
  & round 30 & 0.900 & \textbf{0.096} & 0.99 & \textbf{0.07} \\
  & round 60 & 0.900 & \textbf{0.096} & 0.98 & \textbf{0.13} \\
\bottomrule
\end{tabular}}
\end{minipage}
\end{table}

\paragraph{Impact of attack schedule.}
Table~\ref{tab:attack-schedule-ablation} reports the effect of delaying the
onset of the A3FL backdoor attack (round~0, 30, and 60) on both convergence
and detection quality for M-Krum, NNM, and our R2-NNM. Two trends emerge. First, delaying the attack has a negligible effect on \emph{convergence}:
ACC stays within a narrow band (0.893--0.900) and the
ASR remains low (0.096--0.098) across all schedules and
defenses, indicating that none of the three mechanisms is meaningfully
destabilized by a later attack onset. This is expected, as all three
methods retain a Krum selection step that continues to bound the
malicious clients' influence on the aggregate. Second, and more informative, is the effect on \emph{detection}. TPR is uniformly high (0.98--0.99) regardless of schedule, so malicious clients
are essentially never missed. The interesting variation is in FPR, which
increases monotonically with a later attack start for every defense --- a
later onset gives the attacker more rounds to blend into the benign
update distribution before it is flagged, so legitimate clients are
increasingly misclassified as adversarial. However, the \emph{magnitude}
of this degradation differs sharply between methods. M-Krum's FPR nearly
quadruples, from 0.07 at round~0 to 0.26 at round~60. NNM and R2-NNM are far more robust to this
effect, with FPR rising only from 0.00 to 0.13 over the same range ---
roughly half the degradation observed for M-Krum. Comparing NNM and R2-NNM directly, the two share an essentially identical
FPR trajectory (0.00~$\rightarrow$~0.07~$\rightarrow$~0.13), indicating
R2-NNM does not sacrifice detection
precision relative to plain NNM. Where R2-NNM distinguishes itself is in
attack suppression: it achieves the lowest ASR of the three defenses at
every schedule,
with the gap widening slightly as the attack is delayed. This suggests
that accumulating client reputation over rounds provides an additional signal that continues to suppress the backdoor's
effectiveness. Overall, these results indicate that
R2-NNM improves robustness to attack-timing strategies
without trading off detection accuracy.

\section{Conclusion}
\label{sec:conclusion}
In this paper, we presented \textit{R2CFL}, a robust reputation-driven framework for Crowdsourced Federated Learning (CrowdFL) designed to address the limitations of existing reputation-based approaches under stealthy and adaptive adversaries. While prior work has primarily focused on improving participant selection through reputation mechanisms, it often fails to explicitly quantify or enforce robustness against attackers capable of evading standard detection strategies. To address this limitation, R2CFL tightly integrates a reputation model with a nearest-neighbor mixing defense (R2-NNM), ensuring that reputation evolution is directly coupled with robust update filtering during aggregation. This design prevents stealthy attackers from progressively gaining trust and influencing future learning tasks. More generally, the framework establishes a principled link between defense outcomes and reputation dynamics, allowing reputation scores to faithfully reflect the statistical behavior of the underlying aggregation method.
We note the current design does not adapt to round-varying adversarial pressure, potentially resulting in a suboptimal robustness--fairness trade-off under heterogeneous or non-stationary attacks. We view learning these hyperparameters online, \eg, via a reinforcement-learning controller, as a promising direction for future work.

\medskip

\small \textbf{Acknowledgments.} This research is supported by the CKRISP project (ANR-23-IAS4-0001).

\bibliographystyle{splncs04}
\bibliography{refs}

\appendix

\section*{Appendices}

\subsubsection*{A. Time overhead.} Since aggregation is performed off-chain in both centralized and decentralized settings, we report only the off-chain overhead. 
We evaluate the scalability of different robust aggregation methods by measuring the average aggregation time (in seconds) as the number of clients increases from 20 to 200 on \texttt{FashionMNIST}. Table~\ref{tab:aggregation_time} reports mean $\pm$ standard deviation over multiple runs. We observe significant differences in computational efficiency across methods. Lightweight defenses such as \textit{FLAME} and \textit{R2-NNM} consistently achieve low aggregation times, while methods such as \textit{AutoDFL} exhibits substantially higher computational overhead. This increase is mainly due to their inference-time evaluation of client updates, where each update is explicitly scored, introducing non-negligible per-client overhead. Compared to NNM, \textit{R2-NNM} benefits from a fully vectorized implementation of distance computation and neighborhood aggregation, avoiding Python-level loops and redundant per-client evaluations. In particular, R2-NNM we leverage batched pairwise distance computation and indexed tensor gathering to improve efficiency compared to NNM. Similarly, M-Krum is implemented in a vectorized scoring fashion over flattened updates to reduce the overhead associated with pairwise selection. These implementation differences explain the observed gap, where inference-heavy (\eg, AutoDFL) scales poorly, while vectorized robust aggregation methods maintain strong scalability.

\begin{table}[th]
\centering
\caption{Aggregation time (seconds) under different numbers of clients. Values are reported as mean $\pm$ standard deviation over multiple runs. The best results are shown in \textbf{bold}, while the second-best results are highlighted in \textcolor{brown}{brown}.}
\label{tab:aggregation_time}
\resizebox{0.8\columnwidth}{!}{
\begin{tabular}{lcccc}
\hline
\textbf{Defense} & \textbf{20} & \textbf{50} & \textbf{100} & \textbf{200} \\
\hline
M-Krum &
$0.232 \pm 0.154$ &
$0.592 \pm 0.267$ &
$2.621 \pm 0.867$ &
$8.596 \pm 2.902$ \\

FLAME &
\textbf{0.060 $\pm$ 0.026} &
\textbf{0.109 $\pm$ 0.032} &
\textbf{0.201 $\pm$ 0.030} &
\textbf{0.347 $\pm$ 0.040} \\

DeepSight &
$0.615 \pm 0.198$ &
$1.004 \pm 0.029$ &
$2.038 \pm 0.289$ &
$4.388 \pm 0.635$ \\

NNM+Krum &
$0.376 \pm 0.238$ &
$1.228 \pm 0.214$ &
$6.252 \pm 0.628$ &
$13.758 \pm 1.180$ \\

AutoDFL &
$17.483 \pm 0.158$ &
$51.174 \pm 0.567$ &
$82.781 \pm 0.686$ &
$97.000 \pm 0.687$ \\

SSMTD &
\textcolor{brown}{$0.101 \pm 0.138$} &
\textcolor{brown}{$0.349 \pm 0.299$} &
\textcolor{brown}{$0.377 \pm 0.181$} &
\textcolor{brown}{$0.627 \pm 0.488$} \\

\textbf{R2-NNM} &
$0.328 \pm 0.163$ &
$0.632 \pm 0.036$ &
$1.198 \pm 0.158$ &
$2.259 \pm 0.163$ \\
\hline
\end{tabular}
}
\end{table}

\begin{table}[th]
\small
\begin{minipage}{0.45\textwidth}
\centering
\caption{Impact of $\alpha$ on the final separation gap $\Delta$. The shaded rows indicate the recommended operating range $\alpha\in[0.25,0.75]$.}
\label{tab:alpha-ablation}
\resizebox{0.95\columnwidth}{!}{
\begin{tabular}{c c c c c}
\toprule
& \multicolumn{2}{c}{\texttt{GTSRB}} & \multicolumn{2}{c}{\texttt{FashionMNIST}} \\
\cmidrule(lr){2-3} \cmidrule(lr){4-5}
\textbf{$\alpha$} & \textbf{IID} & \textbf{Non-IID} & \textbf{IID} & \textbf{Non-IID} \\
\midrule
0.00 & 0.705 & 0.724 & 0.710 & 0.723 \\
\rowcolor{gray!12}
0.25 & 0.709 & 0.723 & 0.713 & 0.723 \\
\rowcolor{gray!12}
0.50 & 0.709 & 0.718 & 0.711 & 0.719 \\
\rowcolor{gray!12}
0.75 & 0.711 & 0.715 & 0.715 & 0.716 \\
1.00 & 0.712 & 0.711 & 0.712 & 0.711 \\
\bottomrule
\end{tabular}}
\end{minipage}
\begin{minipage}{0.48\textwidth}
\centering
\caption{Impact of the warm-up length $T_{\text{warm}}$ and $\kappa$ (A3FL, \texttt{GTSRB}, non-IID).}
\label{tab:twarm-kappa-ablation}
\resizebox{0.95\columnwidth}{!}{
\begin{tabular}{c c c c c c}
\toprule
\textbf{$T_{\text{warm}}$} & \textbf{$\kappa$} & \textbf{ACC} & \textbf{ASR} & \textbf{TPR} & \textbf{FPR} \\
\midrule
\multirow{3}{*}{5}
  & 0.25 & 0.920 & 0.041 & 0.96 & 0.29 \\
  & 0.50  & 0.922 & 0.042 & 0.96 & 0.29 \\
  & 1.00    & 0.917 & 0.044 & 0.91 & 0.29 \\
\midrule
\multirow{3}{*}{25}
  & 0.25 & 0.919 & 0.041 & 0.96 & 0.24 \\
  & \textbf{0.50}  & \textbf{0.917} & \textbf{0.041} & \textbf{0.97} & \textbf{0.23} \\
  & 1.00    & 0.917 & 0.043 & 0.92 & 0.25 \\
\midrule
\multirow{3}{*}{50}
  & 0.25 & 0.919 & 0.041 & 0.96 & 0.26 \\
  & 0.50  & 0.918 & 0.042 & 0.97 & 0.23 \\
  & 1.00    & 0.918 & 0.045 & 0.92 & 0.25 \\
\bottomrule
\end{tabular}}
\end{minipage}
\end{table}

\subsubsection*{B. Impact of $\alpha$ on the final $L_{rep}$ gap.} 
Table~\ref{tab:alpha-ablation} sweeps $\alpha$ (Eq.~\eqref{eq:contrib}) over $[0,1]$ and reports the final separation gap $\Delta$ between benign and malicious clients under IID and non-IID partitions. Overall, $\Delta$ remains relatively stable across the considered values of $\alpha$. For \textit{GTSRB}, the IID gap increases from $0.705$ at $\alpha=0$ to $0.712$ at $\alpha=1$, whereas the non-IID gap decreases from $0.724$ to $0.711$. This behavior is consistent with the intuition that, under non-IID data, relying solely on update distances can be less reliable, as benign clients may naturally exhibit greater dispersion due to heterogeneous local data distributions. Incorporating the complementary contribution component therefore helps mitigate the effect of statistical heterogeneity on separation. A similar trend is observed on \textit{FashionMNIST}, where the IID gap remains within $0.710$--$0.715$ and the non-IID gap within $0.711$--$0.723$. The discrepancy between IID and non-IID settings remains small across all $\alpha$ values, with the largest difference observed at $\alpha=0$ ($\approx 0.021$) for \textit{GTSRB}. This indicates that the proposed mixing strategy is relatively robust to data heterogeneity. 
We therefore consider $\alpha\in[0.25,0.75]$ as a suitable operating range. Values in this interval provide a favorable balance between separation performance and robustness to data heterogeneity.

\subsubsection*{C. Impact of the warm-up phase $T_{\text{warm}}$ and $\kappa$.} Table~\ref{tab:twarm-kappa-ablation} investigates the influence of the warm-up duration $T_{\text{warm}}$ (Eq.\eqref{eq:warmup}) $\in \{5,25,50\}$ and the reputation scaling factor $\kappa \in \{0.25,0.5,1.0\}$ on both learning performance (ACC and ASR) and detection effectiveness (TPR and FPR). The results show that a moderate reputation influence combined with a sufficiently long warm-up period provides the best trade-off between robustness and stability. Among the evaluated settings, $(T_{\mathrm{warm}}=25,\kappa=0.5)$ emerges as the most effective configuration, achieving the highest TPR while maintaining one of the lowest FPR values and preserving strong learning performance.

\end{document}